# CatRetriever: Contrastive Representation Learning for Slab-to-Bulk Retrieval in Generative Catalyst Discovery

*Jungho Oh[1,†], Woosung Kim[2,†], Dong Hyeon Mok[3], Jonggeol Na[4,5,6*] and Seoin Back[1,6,7,8*]*

AUTHOR ADDRESS

[1]KU-KIST Graduate School of Converging Science and Technology, Korea University, Seoul 02841, Republic of Korea

[2]Department of Materials Science and Engineering, Korea University, Seoul 02841, Republic of Korea

[3]Department of Chemical and Biomolecular Engineering, Sogang University, Seoul 04107, Republic of Korea

[4]Department of Chemical Engineering and Materials Science, Ewha Womans University, Seoul 03760, Republic of Korea

[5]Department of Chemical Engineering, Graduate Program in System Health Science and Engineering, Ewha Womans University, Seoul, 03760, Republic of Korea

[6]Institute for Multiscale Matter and Systems (IMMS), Ewha Womans University, Seoul 03760, Republic of Korea

[7]Department of Integrative Energy Engineering, Korea University, Seoul 02841, Republic of Korea

[8]Center for Hydrogen and Fuel Cells, Korea Institute of Science and Technology(KIST), Seoul 02792, Republic of Korea

[†]These authors contributed equally to this work.

[*]Corresponding authors: sback@korea.ac.kr (SB), jgna@ewha.ac.kr (JN)

**ABSTRACT**

Inverse design is an emerging data-driven paradigm for efficiently navigating vast chemical spaces to discover new materials with targeted properties, and in the context of heterogeneous catalysis, surface generative models have recently advanced this goal by directly generating catalyst surface-adsorbate structures. However, these models typically operate at the slab level and do not provide the corresponding parent bulk structure, making it difficult to assess bulk-dependent properties such as formation energy, surface energy, crystallographic symmetry, and synthesizability. Here, we address this missing slab-to-bulk connection as a retrieval problem and introduce CatRetriever, a contrastive representation learning model that aligns slab and bulk crystal representations in a shared latent space. From a slab query, CatRetriever accurately retrieves plausible parent bulk candidates with R@1 > 91% and R@3 > 98% on both the in-distribution and holdout evaluation sets. We further extend the CatRetriever framework into an adsorption energy targeted bulk discovery pipeline that combines bulk retrieval, generative search space expansion, and adsorption energy distribution analysis. This workflow evaluates candidates by both structural compatibility with the query slab and their ability to access the target adsorption energy range across diverse surface environments. CatRetriever therefore provides a scalable route for connecting catalyst generative models with physically plausible and adsorption energy compatible bulk catalyst discovery.

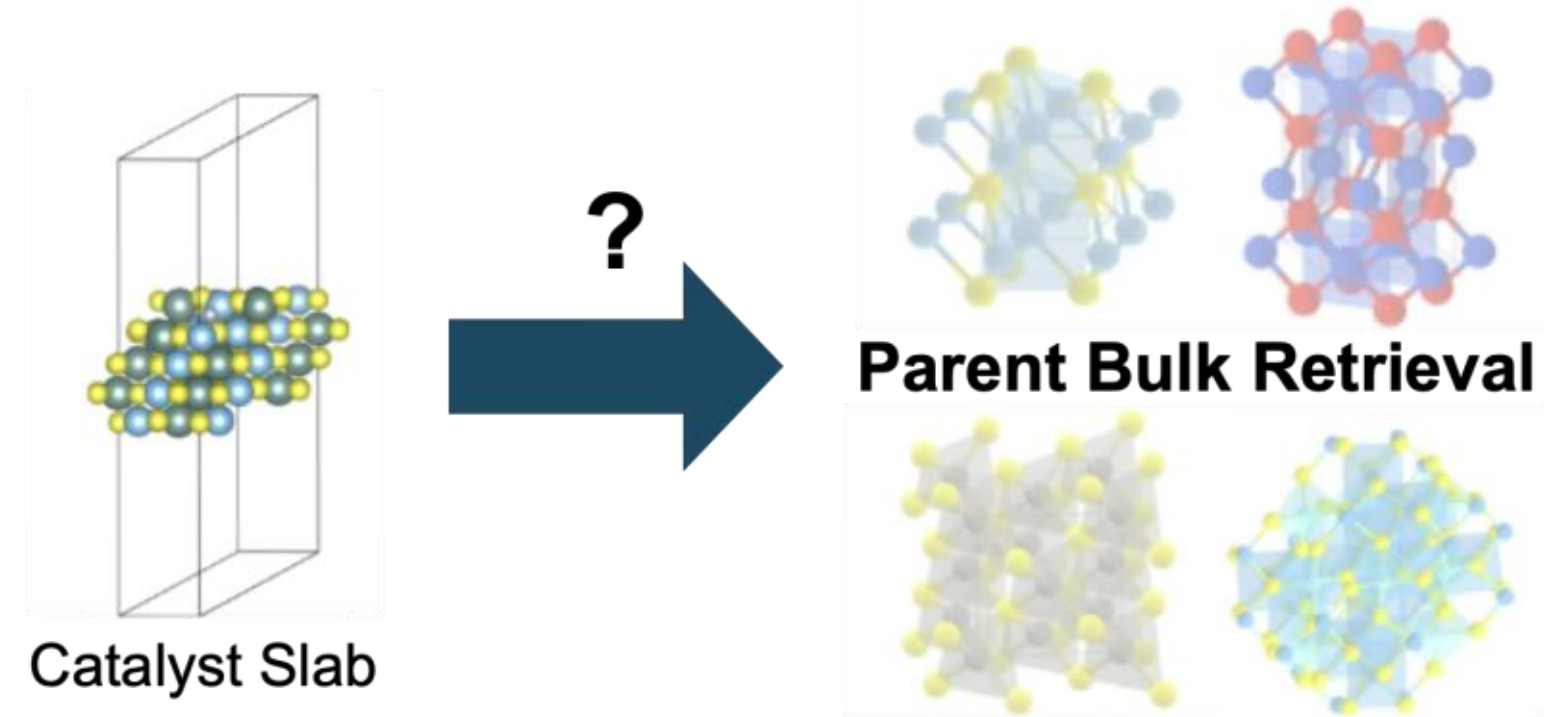

## 1. Introduction

Discovering highly active catalysts is a central goal in enhancing the efficiency of energy conversion processes[1-3]. Traditionally, these efforts have relied on experimental trial-and-error and density functional theory (DFT)-based high-throughput screening, which remain time-consuming and computationally expensive, limiting the breadth of the explorable chemical space. Moreover, conventional screening is inherently constrained by predefined databases or enumerated candidate spaces, and its scalability decreases rapidly as the chemical and configurational complexity of catalyst structures increases[4, 5]. These challenges have motivated a shift from database-bound forward screening toward inverse design strategies. Artificial intelligence (AI) has emerged as a powerful alternative, particularly through generative models capable of proposing novel crystal structures beyond those catalogued in existing databases[6-8].

This generative paradigm has recently been extended to heterogeneous catalytic surfaces[9,10]. Since catalytic reactions occur at surfaces rather than in the bulk, catalyst generative models must account for both surface geometry and adsorbate configurations. The direct generation of slab-adsorbate structures was pioneered by CatGPT, a generative pretrained transformer for heterogeneous catalysts[9]. Subsequent diffusion-based frameworks have advanced adsorbate placement and surface-structure generation[11, 12], while recent flow-matching approaches have enabled the direct co-generation of slab-adsorbate systems[13]. Most recently, conditional generation strategies have enabled the generation of catalyst structures guided by categorical or continuous target properties[14]. Despite these advances, a fundamental limitation persists. Existing surface generative models operate at the slab level and provide no explicit information about the parent bulk structure from which a surface is derived[9, 14]. This gap represents a critical bottleneck in practical catalyst design because essential material descriptors such as formation energy, surface energy, crystallographic symmetry, and space group, are defined with respect to the bulk[15, 16]. Without knowledge of the parent bulk, the thermodynamic plausibility, phase stability, and metastability of a generated surface structure cannot be rigorously evaluated[14, 17]. This challenge has been noted in recent literature, yet it remains largely unresolved and continues to hinder the deployment of generative models in real-world catalyst discovery workflows.

We define this missing connection as the slab-to-bulk retrieval problem: inferring plausible parent bulk structures from a given slab. This task differs from conventional bulk-to-slab enumeration or database lookup, because conditionally generated slabs are not guaranteed to exactly match any pre-enumerated surface. Although direct structural comparison against a slab database is possible, it becomes inefficient for repeated post-generation screening because each query must be compared against a large pool of candidate surfaces. A learned retrieval formulation provides a scalable alternative by embedding slabs and bulks in a shared representation space and ranking candidate bulks through fast similarity search. This formulation also accommodates the non-unique nature of slab-to-bulk inference, where a local surface motif may be compatible with multiple chemically or structurally similar bulk structures, and it can be used as a modular post-generation bridge for existing surface generative models.

To address this problem, we propose CatRetriever, a contrastive representation learning model[18, 19] that captures the structural and chemical correspondence between bulk crystals and catalytic surfaces. CatRetriever achieves approximately 92% top-1 retrieval accuracy on the Materials Project dataset. However, retrieval from a fixed database is inherently limited by the database itself. We therefore develop a two-stage search strategy that augments database retrieval with a crystal structure generative model[8], enabling exploration of bulk candidates beyond the Materials Project database. Building on CatRetriever, we construct an end-to-end catalyst discovery pipeline that integrates conditional catalyst surface generation[14], slab-to-bulk retrieval, and bulk-to-surface enumeration[20] and machine learning force field-based adsorption energy evaluation[21], establishing a physically grounded and scalable workflow for practical catalyst discovery.

## 2. Results and Discussions

We formulate the slab-to-bulk matching problem as a cross-domain retrieval task in a shared embedding space[19]. In CatRetriever, bulk crystals and catalyst surfaces are independently encoded and projected into a common representation space[18], where their compatibility is evaluated through embedding similarity. The shared space is learned contrastively by bringing matched slab-bulk pairs closer together while separating unmatched pairs. Rather than treating parent-bulk identification as a conventional single-label classification problem, CatRetriever returns a ranked list of plausible bulk candidates. This retrieval formulation reflects the inherent non-unique nature of slab-to-bulk inference, as a given surface may be compatible with multiple bulk structures exhibiting similar local chemical and structural environments.

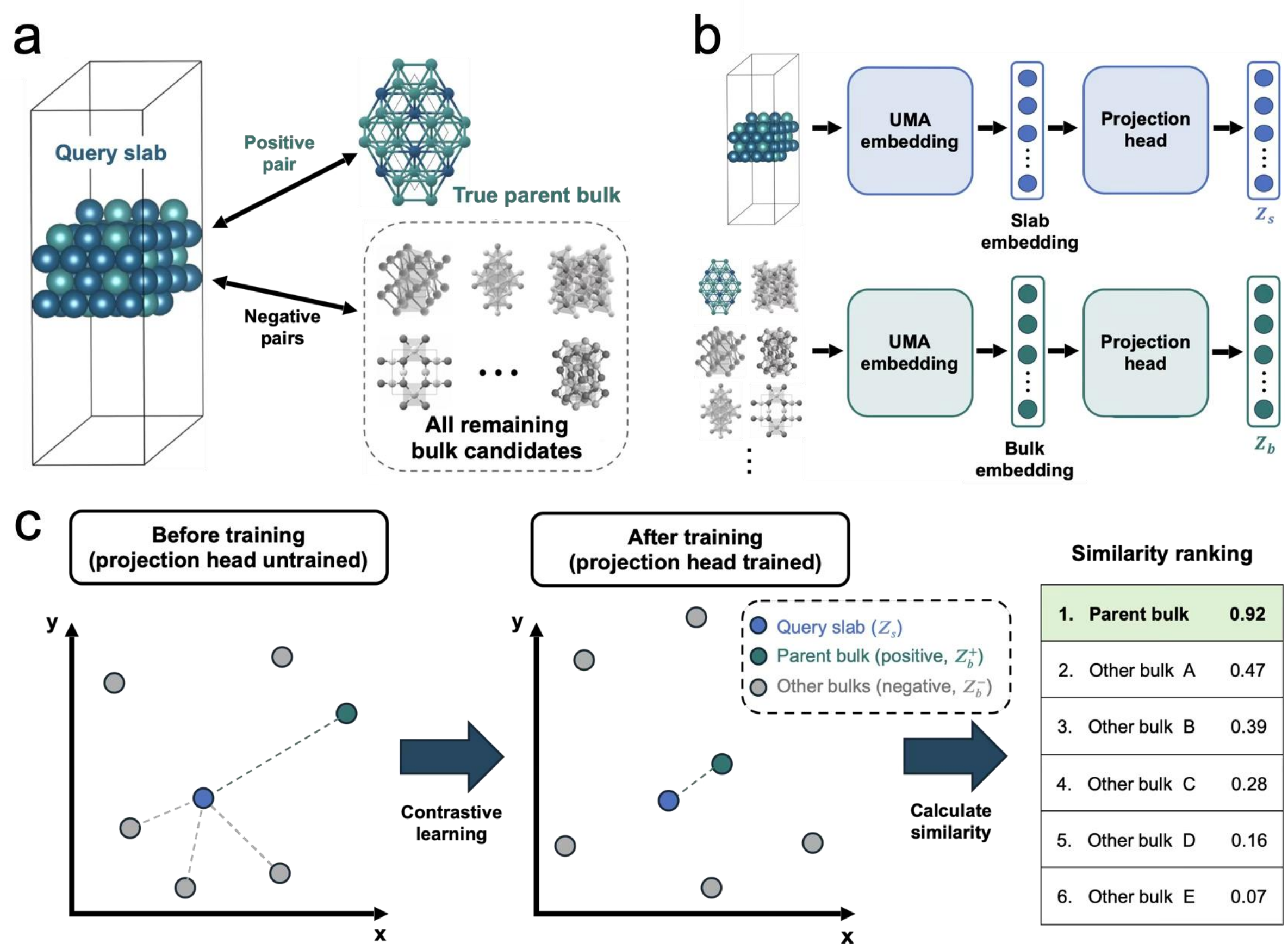

**Figure 1.** Schematic illustration of CatRetriever for contrastive slab-to-bulk retrieval. (a) Definition of positive and negative pairs for contrastive learning. (b) UMA-based embedding extraction and projection of slab and bulk structures into latent representations, $Z_s$ and $Z_b$. (c) Contrastive learning aligns each query slab with its true parent bulk while separating negative bulk candidates, enabling similarity-based parent bulk retrieval.

**Construction of Bulk and Slab Database**

In this work, a bulk candidate database was constructed to define the retrieval search space corresponding to a given slab query. The bulk candidate set was derived from the Materials Project database[15]. Following the compositional space and stability criteria of the Open Catalyst 2020 (OC20) bulk dataset, binary and ternary bulk structures with formation energies below 0 eV/atom and energy above the hull ($E_{hull}$) values below 0.1 eV/atom were selected[20], resulting in a total of 38,901 bulk candidates.

Surface slabs were then generated from these bulks for low-index Miller surfaces using the "Slab.from_bulk_get_specific_millers()" implementation in the FAIRChem/Open Catalyst Project slab generation workflow[20]. This procedure follows established slab-construction methods[22], and enumerates possible surface terminations for each Miller index, allowing multiple slab configurations to be considered for a single parent bulk. Through this process, a slab database of 1,384,140 structures was established for both retrieval training and evaluation. Detailed slab-generation procedures are provided in **Supplementary Note S1**.

**Learning a Joint Slab-Bulk Representation with CatRetriever**

To represent both bulk and slab structures within a shared representation space, structural embeddings were extracted using the pretrained atomic foundation potential, Universal Model for Atoms (UMA; uma-s-1p1)[21]. Each bulk and slab structure was provided as input to the model, and the UMA backbone representations were pooled to obtain fixed-length, structure-level embeddings. The resulting bulk and slab embeddings were then used as input features for CatRetriever, the proposed bulk–slab contrastive retrieval model.

Before training, the bulk–slab dataset was split into model-development and holdout subsets. Specifically, 20% of the bulk structures were first assigned to the holdout set, and all slabs generated from these bulk structures were excluded from training and validation. The remaining 80% of the bulk structures were used for model development, where the corresponding slabs were randomly divided into training, validation, and in-distribution test sets in a ratio of 0.64:0.16:0.20. The resulting in-distribution and holdout test sets were used for the retrieval performance evaluation described in the next section.

To learn the correspondence between slab queries and their parent bulk structures, CatRetriever was trained using a contrastive learning formulation[18, 23]. For each query, the pair with the true bulk was defined as a positive pair, while negatives were formed implicitly using the rest of the full candidate bulk pool, without requiring explicit negative labels (**Figure 1a**). Both bulk and slab embeddings were passed through separate projection heads and mapped into a shared latent space. The resulting latent representations of a slab $s$ and a candidate bulk $b$ are denoted as $Z_s$ and $Z_b$, respectively (**Figure 1b**). The retrieval score was defined as

$$r(s,b) = Z_s^{\top} Z_b$$

which is equivalent to cosine similarity because both representations were L2-normalized. The projection heads were trained with a frozen pretrained backbone of UMA, optimizing the alignment between bulk and slab representations such that a query slab retrieves its corresponding parent bulk with high retrieval score (**Figure 1c**). Details of the UMA embedding extraction procedure, scalar-channel pooling (l = 0), projection-head architecture of CatRetriever, contrastive training, validation, and model selection are provided in **Supplementary Note S2**.

## Parent-Bulk Retrieval Performance

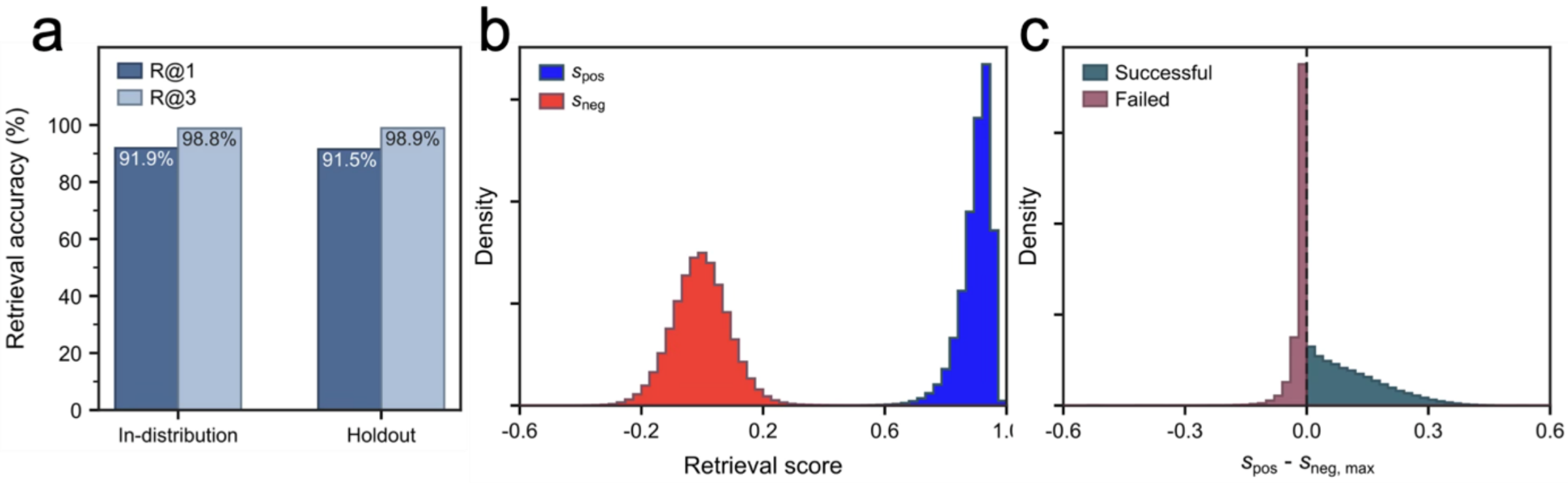

**Figure 2**. Parent-bulk retrieval performance and similarity score distribution of CatRetriever. (a) Retrieval performance on the in-distribution and holdout test sets, evaluated using recall at rank 1 (R@1) and rank 3 (R@3). (b) Distributions of similarity scores for the holdout test set. For visual clarity, the $s_{pos}$ and $s_{neg}$ density distributions were independently rescaled (c) Distributions of the similarity margin for correct and incorrect top-1 retrievals.

We first evaluated how accurately CatRetriever retrieves the corresponding parent bulk from a given slab query. **Figure 2a** summarizes the parent-bulk retrieval performance in terms of Recall@1 (R@1) and Recall@3 (R@3). Here, R@1 denotes the fraction of queries for which the parent bulk is ranked at the top position, while R@3 measures the fraction of queries for which the parent bulk appears within the top three candidates. To assess generalization, the evaluation was conducted on two distinct sets: an in-distribution set and a holdout set. The in-distribution set consists of unseen slab queries derived from bulk structures represented in the training set, enabling evaluation of generalization to new slabs from seen bulks. In contrast, the holdout set comprises slab queries from bulk structures that are entirely excluded from training, enabling evaluation of generalization to unseen bulk structures.

The model achieves strong retrieval performance on both sets. On the in-distribution set, it attains R@1 = 91.9% and R@3 = 98.8%, while on the holdout set, it maintains similarly high performance with R@1 = 91.5% and R@3 = 98.9%. These results indicate that CatRetriever does not merely memorize slab-bulk correspondences observed during training, but generalizes effectively to slab queries derived from previously unseen bulk structures, demonstrating robust out-of-distribution generalization. We note that the objective of this work is not a strict top-1 identification, but reliable shortlisting of a small number of plausible bulk candidates for subsequent adsorption energy evaluation. From this perspective, the consistently high R@3 ($\geq$ 98%) is particularly noteworthy, as it ensures that the correct parent bulk is included within a small candidate set with high probability.

We then examined whether CatRetriever successfully learned a shared representation space that distinguishes matched slab-bulk pairs from non-matching candidates. For each query slab, the similarity to its annotated parent bulk was defined as the positive score, $s_{pos}$, while similarities to non-matching bulk candidates were treated as negative scores, $s_{neg}$. **Figure 2b** shows the corresponding score distributions for the holdout test set. The $s_{pos}$ distribution is markedly shifted toward higher values than the $s_{neg}$ distribution. This clear separation demonstrates that the learned embedding space effectively distinguishes true slab–bulk correspondences from unrelated candidates. Some non-matching bulk structures may share similar local coordination environments, compositions, or structural motifs with the query slab and may therefore receive relatively high similarity scores. Nevertheless, the strong overall separation indicates that the learned similarity provides a meaningful basis for ranking plausible parent-bulk candidates.

To further characterize retrieval failures, we defined the similarity margin as $\Delta s = s_{pos} - s_{neg,\ max}$, where $s_{neg,\ max}$ is the highest similarity among all non-matching bulk candidates. Positive margins indicate that the annotated parent bulk is ranked above every negative candidate, whereas negative margins indicate that at least one incorrect candidate receives a higher similarity score. As shown in **Figure 2c**, correct retrievals exhibit broadly distributed positive margins, while incorrect retrievals are concentrated close to zero on the negative side. This indicates that most retrieval failures occur when the annotated parent bulk and the strongest competing candidate receive very similar scores, rather than when an unrelated bulk is ranked substantially higher. Such low-margin failures likely arise from ambiguous regions of the embedding space, where multiple bulk structures share similar local chemical or structural environments for the query slab. Accordingly, an incorrect parent-bulk assignment under a single-parent annotation does not necessarily imply that the retrieved candidate is chemically implausible. Instead, it may reflect the presence of multiple bulk structures that are similarly compatible with the query surface.

Overall, CatRetriever demonstrates strong capability in recovering relevant bulk candidates from slab queries, particularly in placing the correct bulk within a short candidate list. Collectively, these results suggest that the learned representations effectively capture the underlying bulk-slab correspondence and provide an effective retrieval basis for subsequent adsorption energy-based candidate evaluation.

## Application of CatRetriever-Assisted Bulk Discovery Framework

To demonstrate the practical utility of the CatRetriever-assisted discovery pipeline, we applied it to adsorption energy-driven catalyst discovery for the thermal ammonia decomposition reaction ($2NH_3 \rightarrow N_2 + 3H_2$)[24, 25]. The overall workflow is summarized in **Figure 3**, where target-conditioned slab generation, CatRetriever-based parent-bulk retrieval, MatterGen-based search-space expansion, and adsorption energy validation are integrated into a single discovery pipeline. Here, we used CatGPT, recently extended to support adsorbate and adsorption energy conditioned generation[14]. CatGPT is an autoregressive catalyst generative model that represents slab–adsorbate structures as tokenized sequences[9] and generates them under prescribed categorical and continuous conditions[14]. In this application, we fixed the adsorbate identity to N* and imposed the N* adsorption energy as a continuous target condition of −0.90 eV to generate candidate surface motifs relevant to the thermal ammonia decomposition[26]. Among the CatGPT-generated 10,000 N* adsorbed slab structures, 581 slabs exhibited UMA-calculated N* adsorption energies within the target window of −1.00 to −0.80 eV and were selected as query structures for subsequent parent-bulk retrieval. Details for CatGPT-based generation and screening of adsorption structures are described in **Supplementary Note S3.**

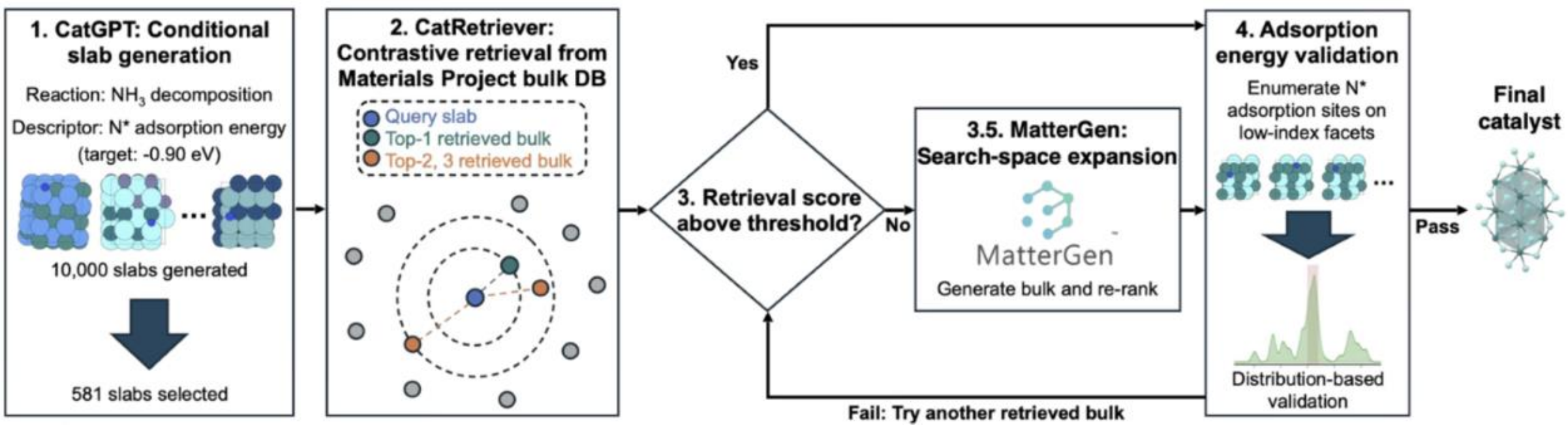

**Figure 3.** Overview of the CatRetriever-Assisted Bulk Discovery Framework. Target-conditioned slab–adsorbate structures are generated using CatGPT and used as queries for CatRetriever-based parent-bulk retrieval. When MP database retrieval is unable to provide bulk structures with high retrieval scores, MatterGen expands the candidate space with compositionally and structurally diverse bulk structures. Adsorption energies on surfaces generated from the retrieved candidates are then evaluated using UMA, and promising catalyst candidates are identified based on the portion of adsorption energies within the target range.

Each selected query slab was used as input to CatRetriever to identify parent bulk structures from the Materials Project database. Among the 581 target-window query slabs, 539 had at least one available MP bulk candidate within the same chemical system, while the remaining 42 had no matching MP entry. We applied a retrieval score threshold of 0.84, above which candidates were treated as structurally plausible parent bulk structures. The rationale for this threshold selection is described in **Supplementary Note S4**. Among the 539 MP-matched cases, 158 exceeded this threshold and were passed to the adsorption energy validation stage.

For the remaining 381 MP-matched cases whose best MP retrieval scores fell below the threshold, as well as the 42 cases with no matching MP entry, we applied MatterGen to expand the search space[8]. The generated bulks were designed to share the same chemical system as the input slab and were incorporated into the existing bulk candidate pool. Details of the MatterGen-based generation procedure are provided in **Supplementary Note S5.1**.

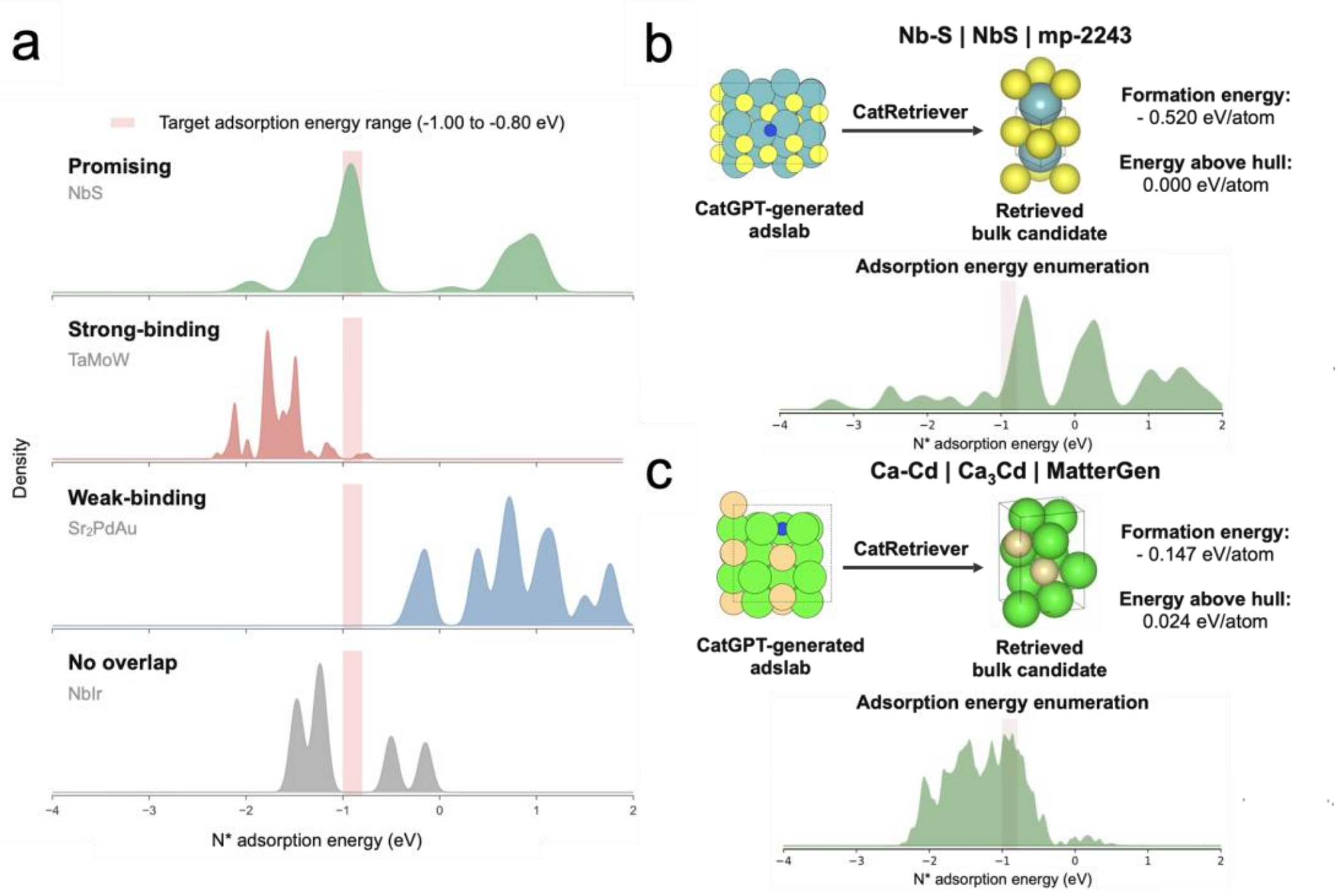

**Figure 4.** (a) Representative N* adsorption energy distributions obtained from bulk candidates with retrieval scores above the threshold. Bulk properties of representative candidates retrieved from (b) the Materials Project database and generated by (c) Mattergen, respectively.

Specifically, 64 additional bulk structures were generated for each corresponding chemical system, and these generated structures were then embedded and re-ranked using CatRetriever. This procedure enabled direct comparison between the best MP candidate and the best MatterGen-generated candidate for each query slab. MatterGen improved upon the best MP retrieval score in 249 out of 381 cases (65%), and 70 of these generated candidates exceeded the retrieval score threshold of 0.84. Representative high-scoring cases, in which the best MatterGen-generated candidates achieved retrieval scores above 0.90 after re-ranking with CatRetriever, are summarized in **Table S1 and S2**. To verify that this improvement did not simply originate from regenerating structures equivalent to the MP-retrieved top-1 candidates, we further performed a StructureMatcher-based redundancy and MP-overlap analysis of the MatterGen-generated candidates[27] (**Supplementary Note S5.2**). The MatterGen and MP top-1 overlap was low, with a mean overlap ratio of 1.16% and a median value of 0%, indicating that the generated candidates generally expanded the search space beyond the existing MP retrieval result. These results demonstrate that even when the existing MP database does not provide a sufficiently compatible parent bulk, generative search-space expansion can recover plausible bulk candidates that are more structurally consistent with the query slab structures.

Based on the retrieval-score analysis presented in the previous section and **Supplementary Note S4**, a high CatRetriever score indicates strong structural compatibility between the generated slab and a candidate parent bulk. Such structural compatibility, however, does not guarantee that the bulk exposes surface environments with the desired N* adsorption strength[3, 26]. Therefore, MP-derived and MatterGen-generated candidates with retrieval scores above the threshold were subjected to adsorption energy validation. For each candidate bulk, a surface corresponding to the Miller index of the generated query slab was first constructed using the slab generation utilities in the FAIRChem/Open Catalyst Project. Possible N* adsorption sites were systematically enumerated, and structurally equivalent adsorption configurations were filtered out to define a representative adsorption set. Adsorption energies were then calculated using uma-s-1p1. This validation can be completed within a relatively short computational time, requiring approximately 26 minutes per candidate bulk using a single GPU (**Supplementary Note S6**).

As summarized in **Figure 3a**, bulk candidates with retrieval scores above the threshold were classified into four representative types according to their N* adsorption energy distributions relative to the target window. (i) Promising candidates exhibited a major adsorption energy density region overlapping the −1.00 to −0.80 eV target window, indicating that the retrieved bulk exposed surface sites with the desired N* binding strength. (ii) Strong binding candidates showed distributions shifted toward more negative adsorption energies, suggesting excessive N* stabilization. (iii) Weak binding candidates showed distributions shifted toward less negative adsorption energies, indicating too weak N* binding[3]. (iv) No overlap candidates exhibited negligible adsorption site density within the target window, without a clear shift toward either strong or weak binding. This distribution-based classification provides a functional validation layer that distinguishes structurally plausible candidates from catalytically relevant ones.

Based on the combined retrieval score and adsorption energy criteria, target-accessible structures were selected as final catalyst candidates. **Figure 4b** and **4c** show representative candidates obtained from the MP database and MatterGen, respectively, namely an MP-retrieved NbS and a MatterGen-generated $Ca_3Cd$. The corresponding CatRetriever scores and query-orientation N* adsorption energy distributions are summarized in **Supplementary Note S7**, confirming that both candidates satisfied the retrieval score threshold and exhibited substantial overlap with the target adsorption energy window.

Importantly, identifying the corresponding bulk structure provides two distinct advantages beyond simply assigning a likely parent phase to the query slab. First, it enables the evaluation of bulk-dependent thermodynamic properties that cannot be inferred from the isolated slab alone. Accordingly, bulk-level thermodynamic properties were evaluated using the appropriate source for each candidate. The formation energy and energy above hull of the MP-retrieved NbS candidate were obtained from the Materials Project database, whereas those of the MatterGen-generated $Ca_3Cd$ candidate were evaluated by DFT calculations and implemented into the MP database to calculate formation energies and the energy above the hull in the consistent way (**Supplementary Note S8**). These results indicated that both candidates are thermodynamically plausible, with NbS located on the convex hull and $Ca_3Cd$ exhibiting a low hull energy of 0.024 eV/atom. Second, the identified bulk serves as a structural source from which multiple crystallographic facets and adsorption sites can be enumerated, allowing the surface-dependent distribution of N* adsorption energies to be evaluated beyond the original query-slab orientation. For this analysis, five low-index Miller indices were used. The resulting N* adsorption energy distributions across enumerated surfaces are presented as histograms in **Figure 4b** and **4c**. Together, these analyses demonstrate that CatRetriever enables adsorption-optimized slab structures to be connected to bulk catalyst candidates whose thermodynamic stability and surface-dependent catalytic behavior can subsequently be assessed. Detailed procedures for slab construction, adsorption site enumeration, redundancy filtering, and adsorption energy evaluation are provided in **Supplementary Note S9**.

Overall, these results demonstrate that the proposed pipeline can systematically connect catalyst surface structures generated by generative models with physically plausible and functionally validated bulk structures. CatRetriever provides rapid parent bulk retrieval of candidate bulk structures for target-conditioned slab queries, MatterGen expands the chemical search space when the MP database is insufficient, and adsorption energy distribution analysis determines whether the retrieved or generated bulk can expose surface sites with the desired N* binding strength. Furthermore, by decoupling structural retrieval, generative search-space expansion, and adsorption energy-based validation, the workflow provides a scalable route for practical catalyst discovery beyond catalyst generative model alone.

Nevertheless, the current implementation has several limitations that point to directions for future development. First, the present workflow is designed as a post-generation retrieval and validation framework, rather than a fully joint generator of bulk–slab pairs. Accordingly, bulk-level properties such as phase stability, synthesizability, and crystal symmetry are not imposed during the initial catalyst surface generation step. Instead, they are evaluated after generated slabs are connected to plausible parent bulk candidates through CatRetriever. This design provides modular compatibility with existing catalyst generative models, but it limits direct inverse design of catalysts under explicit bulk-property constraints. Future extensions could integrate bulk-conditioned catalyst generation or joint bulk–slab generation, allowing target adsorption properties and bulk-level feasibility criteria to be optimized simultaneously. Second, CatRetriever is currently trained and evaluated within the chemical space covered by the Materials Project database. Its generalization to bulk systems with compositions or structural motifs entirely absent from the training distribution remains an open question. Addressing this limitation will require either broader training data coverage or retrieval architectures with stronger compositional generalization.

## 3. Conclusions

In this work, we developed CatRetriever, a contrastive slab-to-bulk retrieval model designed to bridge the gap between generated catalytic surfaces and their plausible parent bulk structures. By mapping slab and bulk structures into a shared representation space, CatRetriever enables candidate bulks to be ranked directly from slab queries without exhaustive pairwise structural comparison. The model achieved strong retrieval performance on both in-distribution and holdout evaluation sets, with R@1 values exceeding 91% and R@3 values approaching 99%, demonstrating robust generalization beyond the training distribution.

Building on CatRetriever, we constructed an adsorption energy-based catalyst discovery pipeline that combines retrieval-guided candidate selection with adsorption energy distribution validation. This two-stage strategy ensures both the structural compatibility of a candidate bulk with the generated surface and its ability to reproduce adsorption environments within the target energy range. This study establishes slab-to-bulk retrieval as an essential complement to catalyst generative models, offering a physically grounded and modular framework for connecting generated surface structures to bulk-verified catalyst candidates.

## Code Availability

The code developed in this work and relevant information can be found in Github (https://github.com/SeoinBack/CatRetriever).

## Acknowledgements

S.B. acknowledges the support from the Carbon Neutral Industrial Strategic Technology Development Program (RS-2023-00261088) funded by the Ministry of Trade, Industry & Energy (MOTIE, Korea), the National Research Foundation of Korea (NRF) grants funded by the Korea government (MSIT) (RS-2025-00513832 and RS-2025-02214715), and the NRF grant funded by the Korea government (MSIT and MOE) (No. RS-2025-16063688), and generous supercomputing time from KISTI. This research was also supported by Korea Basic Science Institute (National Research Facilities and Equipment Center) grant funded by the Ministry of Science and ICT (No. RS-2024-00404602).

## Competing Interest

The authors declare no competing financial interests.

# Supplementary Information for

# "CatRetriever: Contrastive Representation Learning for Slab-to-Bulk Retrieval in Generative Catalyst Discovery"

*Jungho Oh[1,†], Woosung Kim[2,†], Dong Hyeon Mok[3], Jonggeol Na[4,5,6*] and Seoin Back[1,6,7,8*]*

AUTHOR ADDRESS

[1]KU-KIST Graduate School of Converging Science and Technology, Korea University, Seoul 02841, Republic of Korea

[2]Department of Materials Science and Engineering, Korea University, Seoul 02841, Republic of Korea

[3]Department of Chemical and Biomolecular Engineering, Institute of Emergent Materials, Sogang University, Seoul 04107, Republic of Korea

[4]Department of Chemical Engineering and Materials Science, Ewha Womans University, Seoul 03760, Republic of Korea

[5]Department of Chemical Engineering, Graduate Program in System Health Science and Engineering, Ewha Womans University, Seoul, 03760, Republic of Korea

[6]Institute for Multiscale Matter and Systems (IMMS), Ewha Womans University, Seoul 03760, Republic of Korea

[7]Department of Integrative Energy Engineering, Korea University, Seoul 02841, Republic of Korea

[8]Center for Hydrogen and Fuel Cells, Korea Institute of Science and Technology(KIST), Seoul 02792, Republic of Korea

[†]These authors contributed equally to this work.

## Supplementary Note S1. Generation of slab structures from Materials Project bulk crystals

To describe the slab-generation procedure used in this study, we provide a representative workflow for constructing surface slab models from Materials Project bulk structure. While the main text briefly describes the generation of slab structures from each parent bulk, this supplementary note provides the corresponding implementation details and illustrates how multiple slab configurations can be obtained from a single bulk crystal.

As a representative example, the bulk structure with Materials Project ID **mp-862690** was retrieved using the Materials Project API[1]. The obtained pymatgen Structure object[2] was converted into an ASE Atoms object[3] and subsequently used to initialize the Bulk object in the FAIRChem/Open Catalyst Project[4] slab-generation workflow. Slab structures were then generated for the Miller indices used in this study: (100), (110), (111), (210), and (221). For each Miller index, the Slab.from_bulk_get_specific_millers() method was used to enumerate the possible surface terminations associated with the corresponding crystallographic orientation.

This procedure allows a single parent bulk structure to yield multiple slab models, as distinct atomic terminations can be exposed even for the same Miller plane. Consequently, the generated slab database includes variations in both surface orientation and termination-dependent local atomic environments. These slab structures were used as surface-query representations for retrieval model training and evaluation. Representative code for generating slab structures from mp-862690 is provided in **Code S1**. The bulk structure of mp-862690, and the corresponding generated slab structures are shown in **Fig. S1**.

```python
from mp_api.client import MPRester
from pymatgen.io.ase import AseAtomsAdaptor
from fairchem.data.oc.core import Bulk, Slab

MP_API_KEY = "your-api-key-here"
material_id = "mp-862690"

# Miller indices used in this study
selected_millers = [
    (1, 0, 0),
    (1, 1, 0),
    (1, 1, 1),
    (2, 1, 0),
    (2, 2, 1),
]

# Retrieve the bulk structure from the Materials Project database
with MPRester(MP_API_KEY) as mpr:
    pmg_structure = mpr.get_structure_by_material_id(material_id)

# Convert the pymatgen Structure object to an ASE Atoms object
bulk_atoms = AseAtomsAdaptor.get_atoms(pmg_structure)

# Initialize the FAIRChem/Open Catalyst Project Bulk object
bulk = Bulk(
    bulk_atoms=bulk_atoms,
    bulk_src_id=material_id
)

# Generate slabs for each selected Miller index
all_slabs = {}

for miller in selected_millers:
    slabs = Slab.from_bulk_get_specific_millers(
        bulk=bulk,
        specific_millers=miller
    )
    all_slabs[miller] = slabs
    print(f"Miller index {miller}: {len(slabs)} slab terminations generated")

# Save the generated slab structures as CIF files for visualization
for miller, slabs in all_slabs.items():
    h, k, l = miller
    for i, slab in enumerate(slabs):
        slab_atoms = slab.atoms
        slab_atoms.write(f"{material_id}_{h}{k}{l}_termination_{i}.cif")
```

**Code S1.** Representative code for generating slab structures from mp-862690.

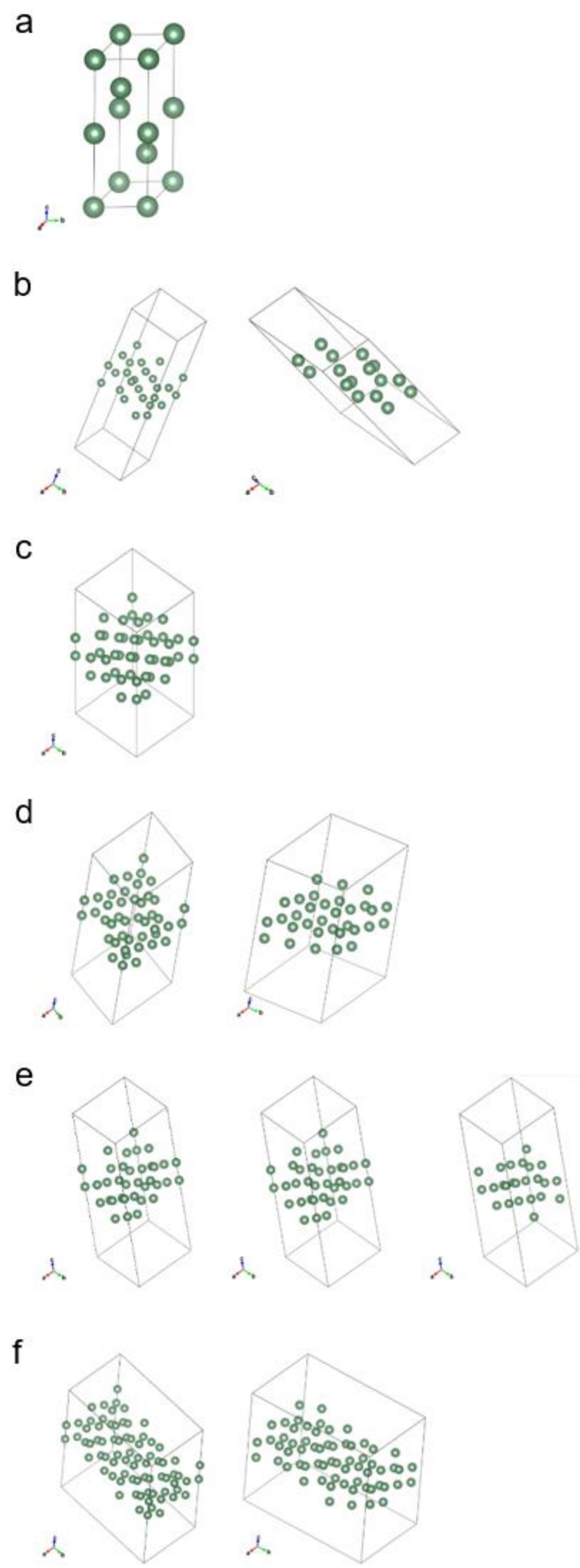

**Figure S1.** The bulk structure of mp-862690 and the generated representative slab structures. (a) bulk structure for mp-862690, generated slabs for (b) (100), (c) (110), (d) (111), (e) (210), (f) (221).

**Supplementary Note S2. UMA-based structure embeddings and contrastive training for CatRetriever.**

## S2.1. Extraction of UMA structure embeddings and scalar-channel pooling

The bulk–slab contrastive learning model was constructed using structure-level embeddings extracted from a pretrained UMA backbone. In this study, we used the UMA-s-1p1 model[5] as the atomistic representation model. During contrastive learning, the UMA backbone was kept frozen, and only the projection heads mapping the bulk and slab embeddings into the shared retrieval space were trained. The procedure used to extract structure-level embeddings from the frozen backbone is described below.

For each input structure containing $N$ atoms, node-level representations were extracted from the frozen UMA backbone by registering a forward hook on the frozen backbone, prior to the trainable projection heads used in the contrastive learning model. The extracted node embedding has a tensor shape of ([$N$, 9, 128]), where $N$ is the number of atoms and 128 is the feature dimension. The second dimension of size 9 originates from the spherical harmonic components associated with the equivariant representation, corresponding to ($l$ = 0, 1, 2) components.

To construct a fixed-length, rotation-invariant structure-level embedding, only the (l = 0) scalar channel was used. Specifically, the first channel of the ([$N$, 9, 128]) node representation was selected, yielding a scalar node representation of shape ([$N$, 128]). Mean pooling was then applied over the atomic dimension to obtain a single ([128])-dimensional vector for each structure. Importantly, the full ([$N$, 9, 128]) tensor was not averaged directly. Instead, only the ($l$ = 0) scalar component was pooled, ensuring that the resulting structure-level embedding was suitable for similarity-based retrieval.

## S2.2. Projection heads and retrieval-score definition

The resulting 128-dimensional bulk and slab embeddings were separately passed through trainable projection heads. Each projection head consisted of a two-layer multilayer perceptron with a hidden dimension of 256: a linear layer from 128 to 256 dimensions, a ReLU activation function, and a final linear layer from 256 to 128 dimensions. The projected bulk and slab embeddings were L2-normalized along the feature dimension before computing dot-product similarity. This normalization constrains each latent vector to have unit length, so that the dot product between two normalized embeddings is equivalent to cosine similarity. As a result, the retrieval score depends on the angular similarity between slab and bulk embeddings rather than on their vector magnitudes, preventing similarity values from being dominated by differences in embedding norm.

### S2.3. Contrastive training, validation, and model selection

CatRetriever was trained using a single-positive InfoNCE[6] objective for slab-to-bulk retrieval. For each mini-batch of $B$ slab queries, the corresponding positive label was defined as the index of the true parent bulk within the candidate bulk pool. Unlike standard in-batch contrastive learning, where only the bulk structures present in the mini-batch serve as negatives, each slab query was compared against the full candidate bulk pool used for training. For a full candidate pool containing $M$ bulk structures, the slab-to-bulk loss was defined as

$$L_{s\to b} = -\frac{1}{B}\sum_{i=1}^{B}\log\frac{\exp\left(sim\left(\boldsymbol{z}_i^s,\ \boldsymbol{z}_{y_i}^b\right)/\tau\right)}{\sum_{m=1}^{M}\exp\left(sim\left(\boldsymbol{z}_i^s,\ \boldsymbol{z}_m^b\right)/\tau\right)}$$

where $\boldsymbol{z}_i^s$ is the normalized latent embedding of the $i$-th slab query, $\boldsymbol{z}_m^b$ is the normalized latent embedding of the $m$-th candidate bulk, $y_i$ is the index of the true parent bulk of the $i$-th slab query, and $\tau$ is the temperature parameter. The similarity function $sim( , )$ was computed as the dot product between normalized embeddings, which is equivalent to cosine similarity. This objective encourages each slab query to retrieve its true parent bulk from the full candidate pool, rather than only distinguishing it from bulk structures within the same mini-batch.

Training was performed with a fixed bulk-level data split (random seed of 0). Of the 38,901 bulk structures, 31,120 were assigned to the training pool, while the remaining 7,781 bulk structures were reserved as a holdout set. Slab–bulk pairs associated with the training-pool bulks were divided into 707,336 training pairs, 177,123 validation pairs, and 220,181 in-distribution test pairs. The bulk-holdout test set contained 279,500 slab–bulk pairs derived from bulk structures that were not included in the training pool.

The model was trained using the AdamW optimizer[7] with a learning rate of 0.001, a batch size of 64, and an InfoNCE temperature of 0.07. Validation retrieval performance was evaluated after each epoch, and validation R@1 was used as the criterion for model selection. Early stopping was applied when the validation R@1 did not improve for 10 consecutive epochs. As shown in **Figure S2**, the training loss decreased steadily during optimization, while the validation retrieval performance improved steadily before reaching a plateau. The best checkpoint was selected at epoch 22, yielding a validation R@1 of 0.9321.

The selected checkpoint was evaluated on both the in-distribution and holdout test sets against the full candidate pool of 38,901 bulk structures. The model achieved R@1 and R@3 values of 0.9195 and 0.9880, respectively, on the in-distribution test set, and 0.9151 and 0.9896, respectively, on the holdout test set.

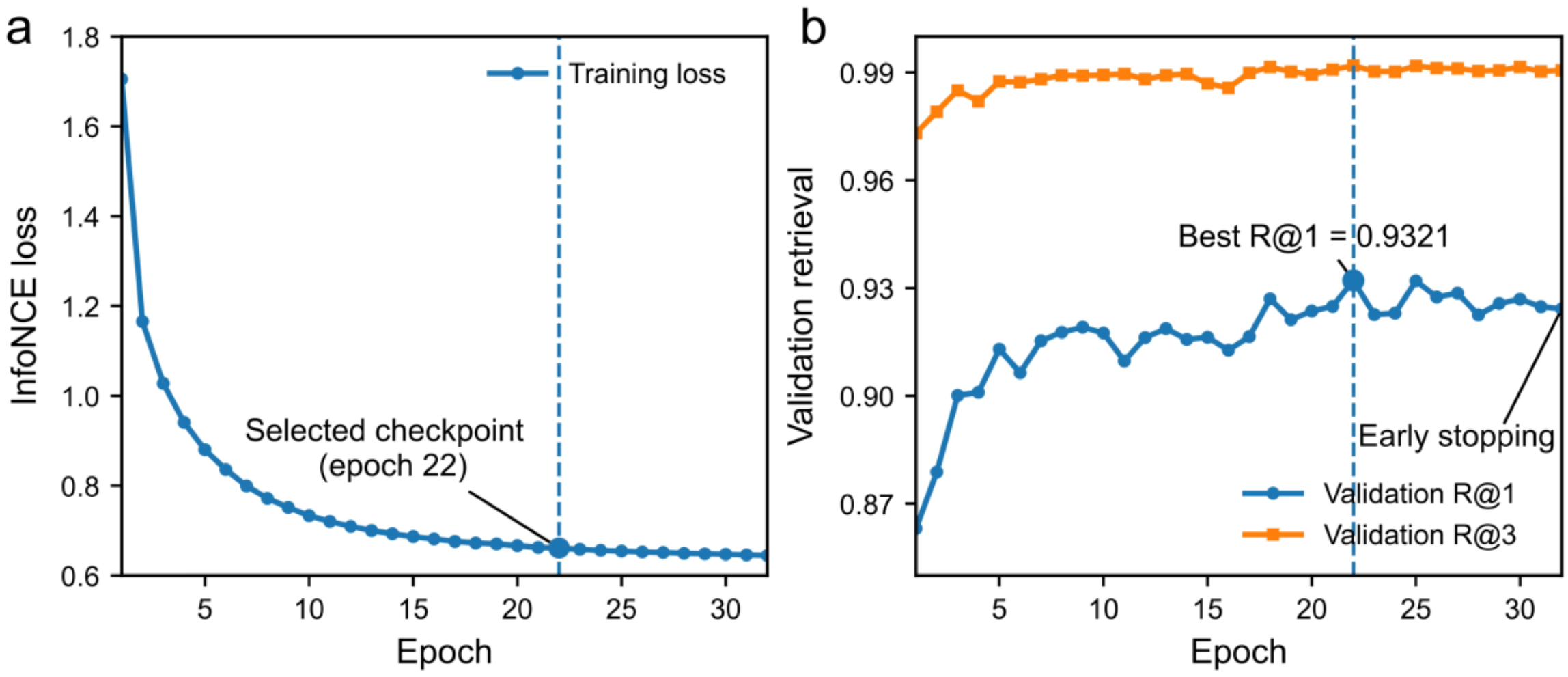

**Figure S2.** Training and validation behavior of CatRetriever. (a) Training InfoNCE loss as a function of epoch. The selected checkpoint was obtained at epoch 22. (b) Validation R@1 and R@3 during training. The final checkpoint was selected based on the highest validation R@1, with early stopping applied after 10 epochs without further improvement.

**Supplementary Note S3. CatGPT-based generation and screening of adsorption**

**structures**

Conditional CatGPT[8] was used to generate N*-adsorbed slab structures under a target adsorption energy condition relevant to thermal ammonia decomposition. The adsorbate token, N, and a target N* adsorption energy of -0.90 eV[9] were provided to the conditional generation model. Stochastic sampling was performed with a maximum sequence length of 1,024 tokens until a total of 10,000 outputs that could be successfully reconstructed as ASE Atoms objects were obtained. Outputs that could not be parsed into valid periodic atomic structures were discarded during generation.

The generated textual representations were converted into periodic adsorption structures using the lattice parameters, slab atomic coordinates, and N* coordinates contained in the model outputs. The generated N atom was assigned a separate atomic tag, allowing it to be distinguished from constituent N atoms in the slab during subsequent structure processing and energy evaluation.

Adsorption energies were evaluated using the UMA-s-1p1 model following the energy definition and convergence criterion described in **Supplementary Note S4**. For each CatGPT-generated adsorption structure, the corresponding clean slab was constructed by removing the tagged N adsorbate. The N-adsorbed slab and the corresponding clean slab were independently relaxed using the BFGS optimizer for up to 200 optimization steps with a maximum-force convergence criterion of 0.05 eV $Å^{-1}$. A generated structure was included in the adsorption energy analysis only if both the adsorbed and clean slab structures satisfied the convergence criterion.

As shown in **Figure S4**, 581 of the 10,000 generated structures exhibited UMA-evaluated N* adsorption energies within the target window of -1.00 to -0.80 eV, corresponding to 5.81% of the generated structures. These structures were retained as query slabs for subsequent parent-bulk retrieval.

For the retrieval analysis, the tagged N adsorbate was removed from each selected structure. A structure-level embedding was then extracted from the resulting adsorbate-free slab using the UMA-based embedding procedure described in **Supplementary Note S2**.

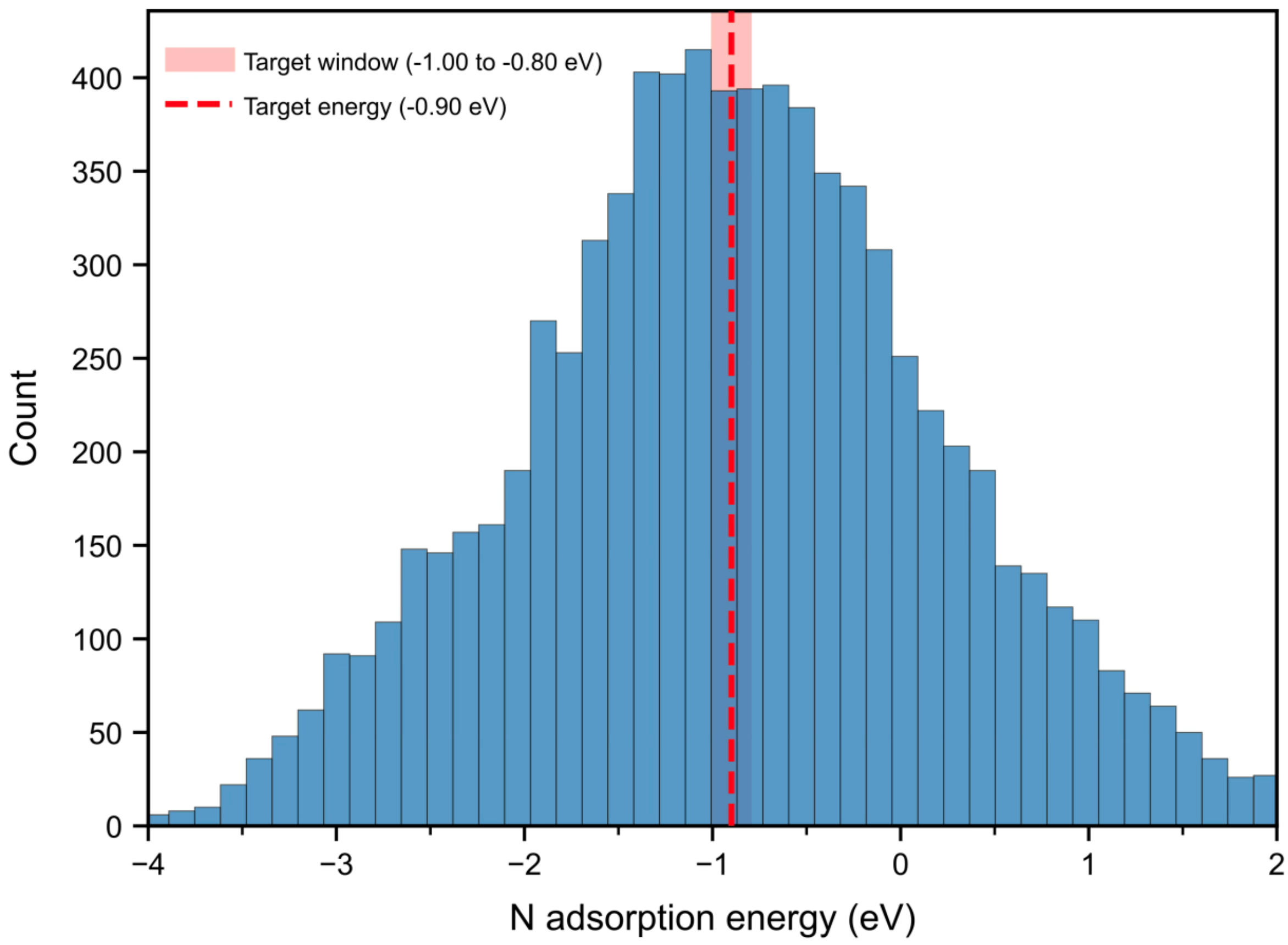

**Figure S3.** Distribution of UMA-evaluated N* adsorption energies for the CatGPT-generated adsorption structures. Among the 10,000 generated structures, 581 structures exhibited adsorption energies within the target window and were selected for subsequent parent-bulk retrieval.

**Supplementary Note S4. Selection of the retrieval-score threshold**

For application to previously unseen slab structures, the true parent bulk is not available, and the top-1 retrieval score must therefore be interpreted as a confidence measure for the predicted slab–bulk correspondence. An operational score threshold was introduced to distinguish plausible slab–bulk matches from low-confidence retrievals before proceeding to downstream candidate evaluation.

For each slab query in the evaluation set, the highest-scoring bulk candidate was regarded as the top-1 prediction. A retrieval was considered correct if the top-1 prediction corresponded to the true parent bulk. At each score threshold, only predictions with a top-1 retrieval score equal to or greater than the threshold were accepted.

Precision was defined as the fraction of accepted predictions for which the top-ranked bulk candidate was correct. Recall was defined as the fraction of all correctly retrieved top-1 predictions that remained after applying the score threshold. Thus, precision measures the reliability of the accepted predictions, whereas recall measures how many of the correctly retrieved cases are retained after thresholding.

**Figure S5** shows the variation in precision and recall as a function of the minimum accepted top-1 retrieval score. Precision remained above 0.90 with only limited variation over the relevant threshold range. In contrast, recall decreased more noticeably as the threshold increased, as some correctly retrieved predictions were excluded due to their lower retrieval scores.

Because the subsequent catalyst discovery workflow was intended to retain most plausible bulk candidates while excluding low-confidence matches, recall was used as the primary criterion for threshold selection. A target recall of 0.90 was selected, and the operational threshold was defined as the highest score threshold that retained at least 90% of all correctly retrieved top-1 predictions.

This procedure resulted in a top-1 retrieval-score threshold of 0.84. At this threshold, 90% of the correctly retrieved cases were retained, while 93% of the accepted predictions corresponded to the correct parent bulk. Therefore, slab–bulk matches with a top-1 retrieval score of 0.84 or higher were regarded as plausible matches in the subsequent catalyst discovery analysis.

The selected threshold should be interpreted as an application-dependent operating point rather than a universal boundary separating correct and incorrect retrievals. A higher threshold may be preferred when reducing false-positive assignments is prioritized, whereas a lower threshold may be used when retaining a broader range of potentially relevant candidates is more important.

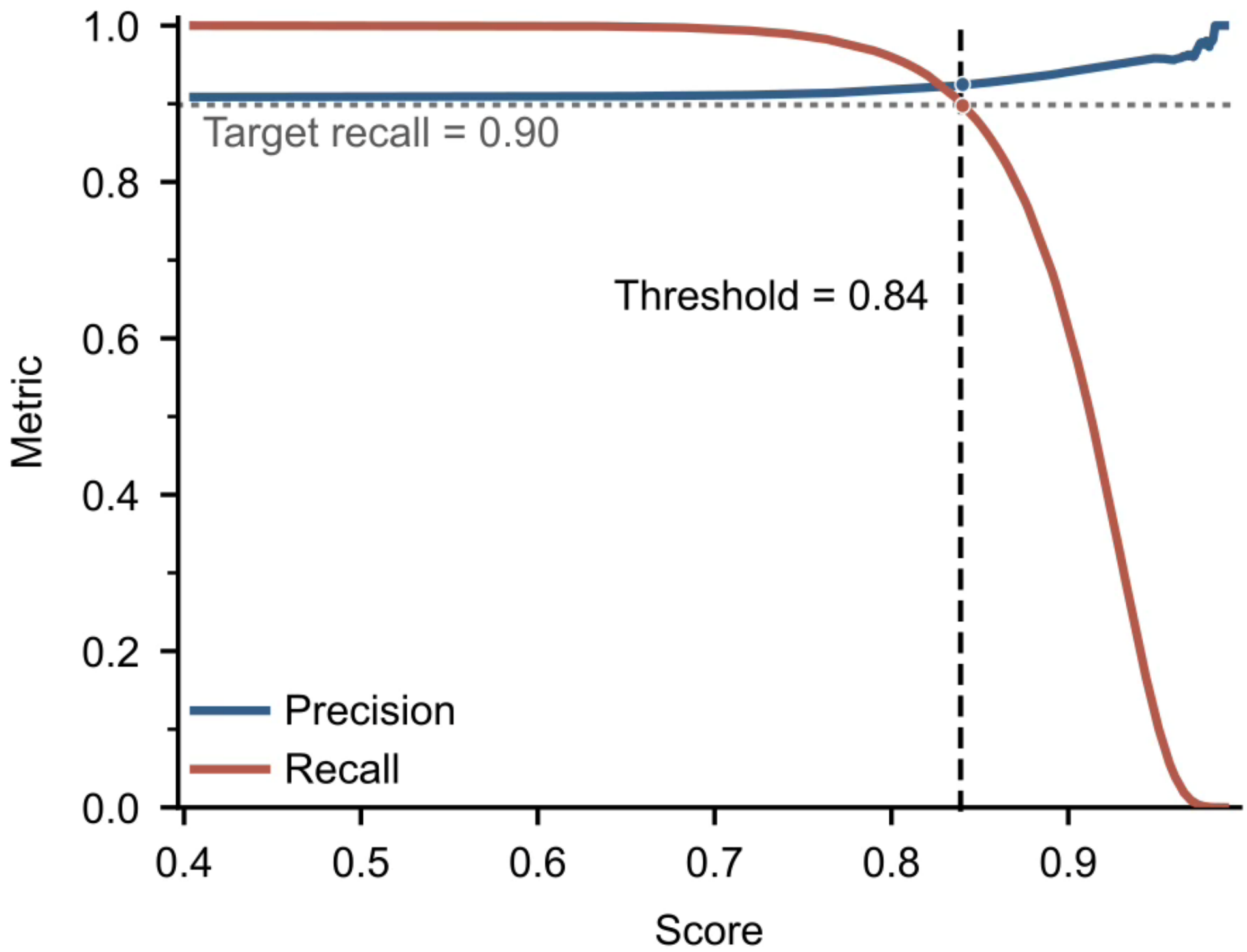

**Figure S4.** Selection of the retrieval-score threshold. Precision and recall as a function of the minimum accepted top-1 retrieval score. The horizontal dotted line indicates the target recall of 0.90, and the vertical dashed line indicates the selected threshold of 0.84. At this threshold, the precision and recall were 0.93 and 0.90, respectively.

**Supplementary Note S5. MatterGen-based bulk generation and post-generation analysis expansion of the bulk candidate search space**

**S5.1. Conditional generation of bulk candidates using MatterGen**

To expand the bulk candidate search space beyond the Materials Project-derived database, we used MatterGen[10] to generate additional bulk structures under property-conditioned generation. This step was introduced for discovery scenarios in which database-based retrieval alone may not provide structurally or compositionally suitable bulk candidates for a given slab query. The generated structures were incorporated into the existing bulk candidate pool and subsequently evaluated through the same slab-to-bulk retrieval workflow.

In this study, MatterGen bulk generation was conditioned on both the chemical system and the energy above the hull. The chemical system was set to match that of the input slab query, thereby constraining the generated structures to contain the same elemental components as the target surface. The energy above the hull was set to 0.1 eV/atom, consistent with the stability criterion used for constructing the Materials Project-derived bulk candidate database. We used the pretrained MatterGen checkpoint "chemical_system_energy_above_hull", which supports joint conditioning on chemical system and energy above the hull.

A representative command used for MatterGen-based bulk generation is shown in **Code S2**. For a target chemical system, Al-Ti, candidate bulk structures were generated with a batch size of 64 and one generation batch. The diffusion guidance factor was set to 2.0 to guide the generation process toward the specified property conditions. The generated bulk structures were then added to the candidate bulk pool, embedded using the same UMA-based embedding procedure, and ranked against the slab query using the trained slab-to-bulk retrieval model. The resulting bulk structures are shown in **Figure S3**.

**S5.2. Redundancy and MP-overlap analysis of MatterGen-generated bulk candidates**

Because MatterGen was used here as an external candidate-generation module, we did not attempt to evaluate the full generative performance of MatterGen. Instead, we performed a lightweight post-generation analysis to determine whether the MatterGen-generated bulk candidates used in our workflow were redundant with one another or overlapped with the MP-retrieved top-1 bulk candidates. This analysis was intended to assess whether the MatterGen step expanded the candidate pool beyond the existing MP-retrieval result, rather than to benchmark MatterGen as a generative model.

The analysis was performed for low-scoring query cases, defined as query slabs whose best MP retrieval score was below 0.84. For each such query, we compared the MP-retrieved top-1 bulk with the corresponding MatterGen-generated bulk candidates produced under the same chemical system condition. Structural equivalence was evaluated using the `StructureMatcher` implemented in pymatgen[2] with the following parameters:

```
ltol = 0.2
stol = 0.3
angle_tol = 5
primitive_cell = True
scale = True
attempt_supercell = True
allow_subset = False
```

For each query-level candidate set, the MP-retrieved top-1 bulk and the MatterGen-generated bulk candidates were partitioned into StructureMatcher-unique groups. We computed three redundancy metrics. First, the total duplicate ratio was defined as $\frac{N_{total}-N_{unique}}{N_{total}}$, where $N_{total}$ is the total number of structures in the combined MP + MatterGen candidate set and $N_{unique}$ is the number of StructureMatcher-unique groups. Second, the MatterGen internal duplicate ratio was computed using only the MatterGen-generated candidates for $N_{total}$ . Third, the MP-overlap ratio was defined as the fraction of MatterGen-generated candidates assigned to the same equivalence group as the MP-retrieved top-1 bulk.

All 381 low-confidence query cases were successfully analyzed. Each query-level candidate set contained 64 MatterGen-generated candidates. The mean total duplicate ratio in the combined MP + MatterGen candidate pool was 18.8%, with a median value of 15.4%. The MatterGen internal duplicate ratio was similar, with a mean of 18.5% and a median of 15.6%, indicating that most duplicate structures originated from redundancy within the MatterGen-generated candidate set. In contrast, the overlap between MatterGen-generated candidates and the MP-retrieved top-1 bulk was low, with a mean MP-overlap ratio of 1.16% and a median value of 0%. These results indicate that the MatterGen-generated candidates generally expanded the candidate space beyond the MP-retrieval result, rather than simply reproducing the same bulk structure. Redundancy and MP-overlap statistics for MatterGen-generated candidates are shown in **Table S3**.

```
export MODEL_NAME=chemical_system_energy_above_hull
export SYSTEM=Al-Ti
export RESULTS_PATH=result/$SYSTEM

mattergen-generate "$RESULTS_PATH" \
  --pretrained-name="$MODEL_NAME" \
  --batch_size=64 \
  --num_batches=1 \
  --properties_to_condition_on="{'energy_above_hull': 0.1,
   'chemical_system': \\"$SYSTEM\"}" \
  --diffusion_guidance_factor=2.0
```

**Code S2.** Representative MatterGen command for chemical-system- and stability-conditioned bulk generation.

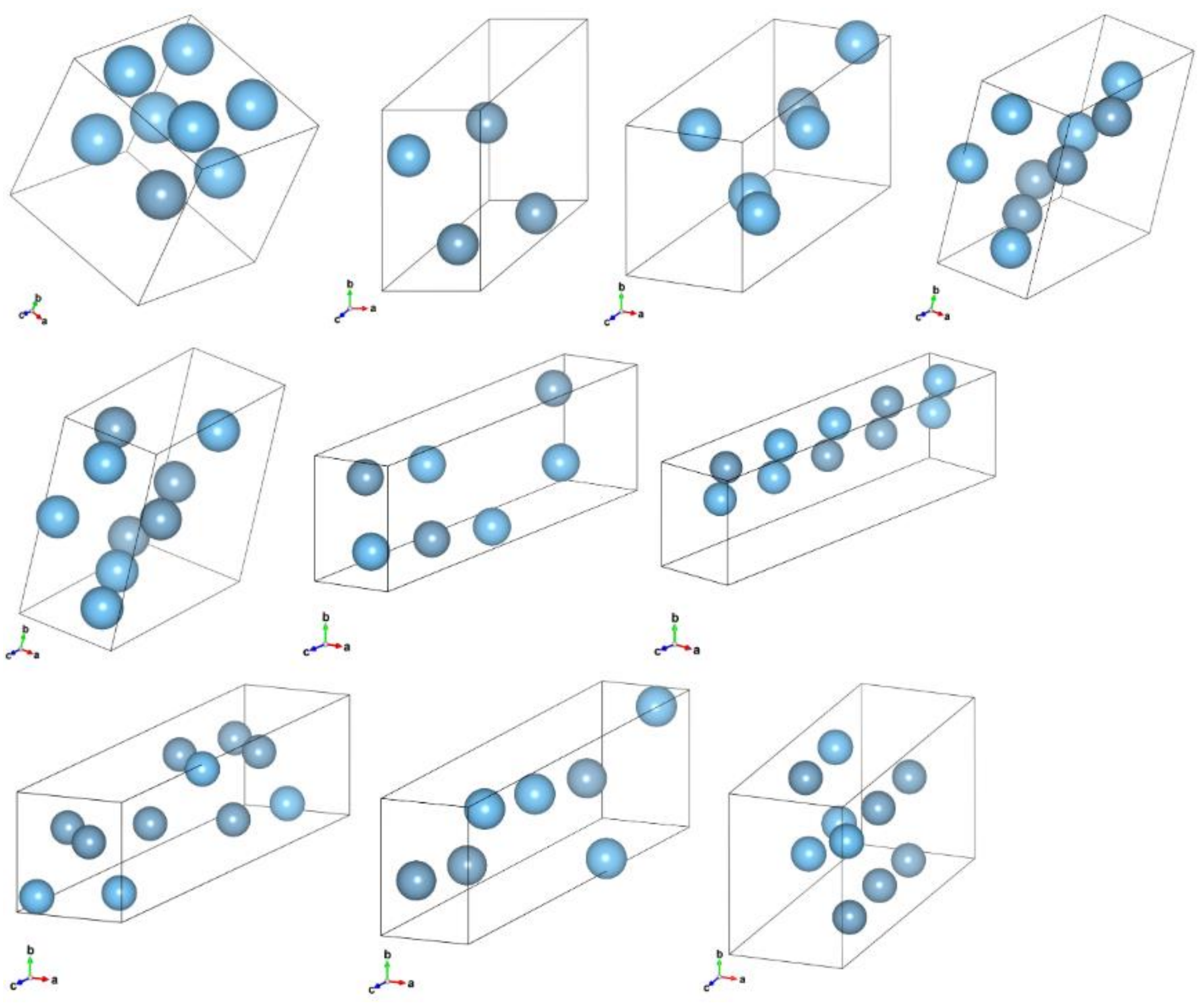

**Figure S3.** Representative bulk structures of a target chemical system, Al-Ti, generated by MatterGen for search space expansion.

**Table S1.** Representative high-scoring MatterGen-generated bulk candidates for low-confidence MP retrieval cases.

| **Query adslab chemical system** | **MP-retrieved top-1 bulk** | **MP-retrieved top-1 bulk score** | **MatterGen-generated top-1 bulk score** |
|---|---|---|---|
| Ca-Zn | mp-2786 | 0.799 | 0.955 |
| Al-Fe-Ge | mp-1183187 | 0.798 | 0.933 |
| Ca-In-Zn | mp-1183608 | 0.797 | 0.916 |
| Al-Co-Hf | mp-5221 | 0.796 | 0.915 |
| In-Zr | mp-20800 | 0.795 | 0.906 |
| Ca-Cd | mp-1183501 | 0.795 | 0.906 |

| Hf-Pd-Re | mp-1184592 | 0.795 | 0.904 |
|---|---|---|---|

**Table S2.** Structural comparison of representative high-scoring cases listed in Table S1.

| Chemical system | Query slab | MP-retrieved top-1 bulk | MatterGen-generated top-1 bulk |
|---|---|---|---|
| Ca-Zn | | | |
| Al-Fe-Ge | | | |
| Ca-In-Zn | | | |
| Al-Co-Hf | | | |
| In-Zr | | | |
| Ca-Cd | | | |

| Hf-Pd-Re | 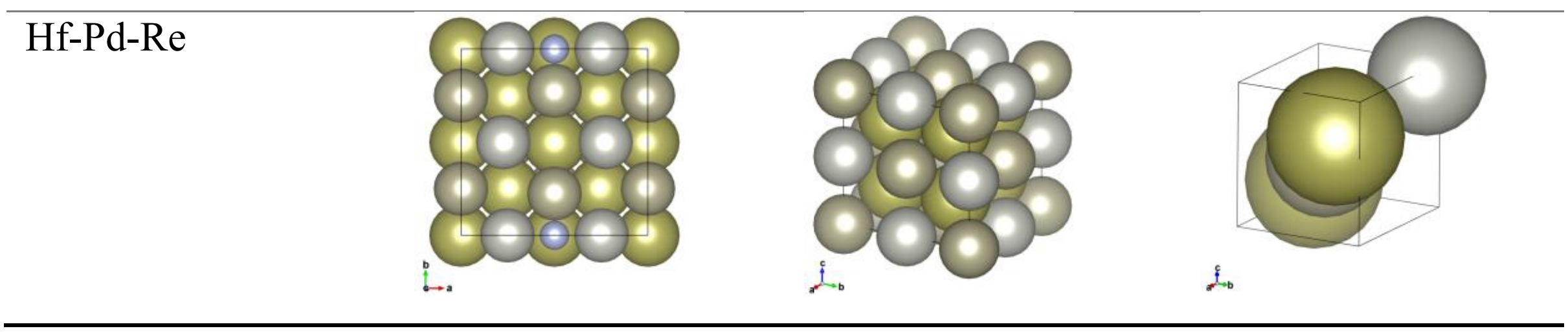 |
| --- | --- |

**Table S3.** Redundancy and MP-overlap statistics for MatterGen-generated bulk candidates.

| **Metric** | **Value** |
| --- | --- |
| Mean total duplicate ratio in MP + MatterGen pool | 18.8% |
| Median total duplicate ratio in MP + MatterGen pool | 15.4% |
| Mean MatterGen internal duplicate ratio | 18.5% |
| Median MatterGen internal duplicate ratio | 15.6% |
| Mean MatterGen–MP top-1 overlap ratio | 1.16% |
| Median MatterGen–MP top-1 overlap ratio | 0.0% |

**Supplementary Note S6. Computational resources**

All model training, evaluation, and adsorption energy calculations were performed using one CPU core of an Intel Xeon Platinum 8452Y processor and a single NVIDIA RTX A6000 GPU with 48 GB of memory.

**Supplementary Note S7. Additional results for the catalyst discovery application**

The representative MP-retrieved NbS and MatterGen-generated $Ca_3Cd$ candidates shown in **Figures 3c and 3d** were selected through the sequential screening procedure used in the catalyst-discovery application. Both candidates first exceeded the CatRetriever retrieval-score threshold of 0.84. Query-slab-oriented adsorption energy validation was subsequently performed by constructing surfaces with the same Miller indices as the corresponding CatGPT-generated query slabs and evaluating their N* adsorption energy distributions following the procedure described in **Supplementary Note S4**. As summarized in **Table S4**, both candidates exhibited adsorption environments within the target binding energy window of −1.00 to −0.80 eV. Because they satisfied both the retrieval-score and query-orientation adsorption energy criteria, their bulk-level properties were further examined and presented as representative application cases in **Figures 3c and 3d**.

**Table S4.** Retrieval scores and query-orientation adsorption energy validation results for the representative NbS and $Ca_3Cd$ application cases.

| | **Retrieval score** | **Query-slab-oriented adsorption energy validation** |
|---|---|---|
| **NbS (MP-retrieved bulk candidate)** | **0.920** | Density; N* adsorption energy (eV); −4 −3 −2 −1 0 1 2 |
| **$Ca_3Cd$ (MatterGen-generated bulk candidate)** | **0.906** | 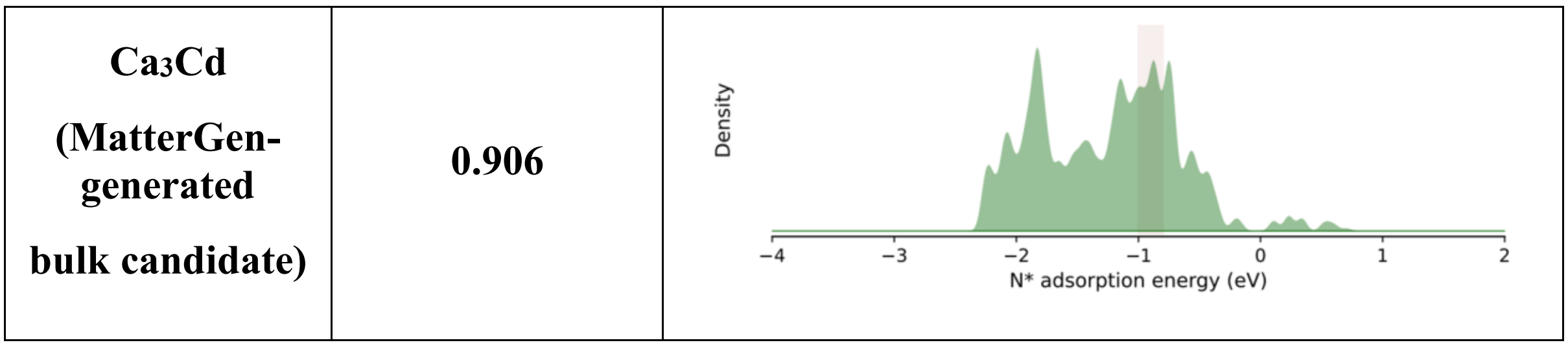  |

**Supplementary Note S8. DFT evaluation**

The generated unit cell contained nine Ca atoms and three Cd atoms, corresponding to three formula units of $Ca_3Cd$. Because the MatterGen-generated $Ca_3Cd$ structure was not available in the Materials Project database, its formation energy was evaluated using an MP-compatible density functional theory workflow. All calculations were performed using VASP (vasp 6.4.1)[11] through the Atomic Simulation Environment (ASE; version 3.26.0), with input parameters generated using pymatgen (version 2024.8.9). The Perdew–Burke–Ernzerhof (PBE) exchange–correlation functional[12] and the Materials Project-compatible projector-augmented-wave potentials[13] were employed. The INCAR parameters and k-point meshes were generated using the `MPMetalRelaxSet` class implemented in pymatgen. To remain consistent with the Materials Project relaxation protocol, structural optimization was performed sequentially in three stages: an initial loose cell optimization with `ISIF = 7`, followed by two full structural optimizations with `ISIF = 3`. In the initial relaxation, a plane-wave cutoff energy of 350 eV was used, and the electronic convergence threshold was set to ten times the corresponding `MPMetalRelaxSet` value. The subsequent two relaxations employed the cutoff energy, electronic convergence threshold, smearing parameters, spin settings, and ionic relaxation parameters generated by `MPMetalRelaxSet`. The $Ca_9Cd_3$ structure and the elemental Ca and Cd reference structures were all calculated using the same three-stage MP-compatible workflow to ensure consistency in the formation-energy evaluation.

The calculated total energies of $Ca_9Cd_3$, elemental Ca, and elemental Cd were

−22.526629, −2.001918, and −0.914590 eV, respectively. The formation energy per atom was calculated as

$$\Delta E_f(Ca_3Cd) = \frac{E_{DFT}(Ca_9Cd_3) - 9E_{DFT}(Ca) - 3E_{DFT}(Cd)}{12},$$

yielding

$$\Delta E_f(Ca_3Cd) = -0.147133 \text{ eV/atom.}$$

The energy above hull was subsequently evaluated using the Ca–Cd phase diagram provided by the Materials Project, accessed on 25 June, 2026. At the $Ca_3Cd$ composition, corresponding to a Cd atomic fraction of 0.25, the convex hull is defined by elemental Ca (mp-45, $\Delta E_f = 0$ eV/atom) and the stable $Ca_3Cd_2$ phase (mp-18167, $\Delta E_f = -0.273$ eV/atom). Linear interpolation between these phases gives a hull energy of

$$E_{hull,\,line} = \frac{0.25}{0.40}(-0.273) = -0.170625 \text{ eV/atom.}$$

Accordingly, the energy above hull of the generated $Ca_3Cd$ structure was calculated as

$$E_{above\ hull} = -0.147133 - (-0.170625) = 0.023492 \text{ eV/atom,}$$

which was rounded to 0.024 eV/atom. This low energy above hull indicates that the MatterGen-generated $Ca_3Cd$ structure is close to the thermodynamic convex hull and is therefore a thermodynamically plausible metastable candidate.

**Supplementary Note S9. Adsorption-configuration generation, redundancy filtering, and adsorption energy evaluation**

Surface slabs were constructed using the FAIRChem/Open Catalyst Project workflow described in **Supplementary Note S1**. For the downstream screening of bulk candidates, query-oriented adsorption energy validation was performed to enable direct evaluation of the query-oriented surface. For the selected final candidates, multi-orientation adsorption energy analysis was performed. Here, the surface space was expanded to the (100), (110), (111), (210), and (221) facets. All slab terminations generated for each crystallographic orientation were included to evaluate the adsorption energy distribution of each bulk candidate over a broader range of possible surfaces.

For each slab, N* adsorption configurations were generated using the `heuristic` and `random_site_heuristic_placement` modes implemented in `AdsorbateSlabConfig`[4]. The configurations identified by the heuristic placement procedure were supplemented with 100 random-site heuristic placements per slab to sample additional local adsorption environments. The configurations generated by the two placement procedures were combined before redundancy filtering.

Because the heuristic and random-site placement procedures could generate repeated configurations, graph-based redundancy filtering was performed before structural relaxation[14]. Each adsorbate–slab configuration was represented as a graph describing the local chemical environment surrounding the adsorbate. Atomic connectivity was determined using an ASE neighbor list based on element-dependent natural cutoff radii, with a cutoff multiplier of 1.1 and a skin distance of 0.25 Å. Periodic connectivity parallel to the slab surface was included during graph construction.

Atoms and atomic connections were represented as graph nodes and edges, respectively. Adsorbate and surface atoms were distinguished through node attributes, while the elemental identities of connected atomic pairs were retained as edge labels. Two configurations were considered redundant when their labeled local-environment graphs were isomorphic under the defined node- and edge-matching criteria. For each group of matching configurations, only one representative structure was retained for subsequent relaxation and energy evaluation.

The clean slab and each retained adsorbate–slab configuration were independently relaxed using the pretrained UMA-s-1p1 model with the OC20 task setting. Structural optimization was performed using an ASE quasi-Newton optimizer for up to 200 optimization steps, with a maximum-force convergence criterion of 0.05 eV $Å^{-1}$.

The N* adsorption energy was calculated as the difference between the energy of the relaxed N-adsorbed slab and the energy of the corresponding relaxed clean slab, with the atomic reference energy of nitrogen subtracted. Here, the energy of the N-adsorbed slab refers to the total energy after relaxation with the adsorbed nitrogen atom, and the clean slab energy corresponds to the relaxed slab without the adsorbate. The nitrogen reference energy was set to −8.083 eV according to the energy-reference convention used for the OC20 task.

The adsorption energies obtained from the retained configurations were aggregated to construct an adsorption energy distribution for each bulk candidate. The same slab construction, adsorbate placement, redundancy filtering, and adsorption energy evaluation procedures were applied to both Materials Project-retrieved and MatterGen-generated bulk candidates. These candidate types differed only in the source from which their initial bulk structures were obtained.